\documentclass[runningheads]{llncs}
\usepackage[T1]{fontenc}
\usepackage{graphicx}
\usepackage{mathrsfs}
\usepackage{multirow}%
\usepackage{amsmath,amssymb,amsfonts}%
\usepackage[title]{appendix}%
\usepackage{xcolor}%
\usepackage{textcomp}%
\usepackage{manyfoot}%
\usepackage{booktabs}%
\usepackage{algorithm}%
\usepackage{algorithmicx}%
\usepackage{algpseudocode}%
\usepackage{listings}%

\usepackage[utf8]{inputenc}
\usepackage[T1]{fontenc}

\usepackage{graphicx}
\usepackage{svg}
\usepackage{tikz}

\usepackage{amsmath,amssymb,amsfonts}%

\usepackage{algorithm}%
\usepackage{algorithmicx}%
\usepackage{algpseudocode}%
\usepackage{cleveref}
\usepackage{pbox}
\usepackage{makecell}
\usepackage{soul}
\usepackage{xcolor}

\usepackage[numbers]{natbib}

\begin{document}
\pagestyle{empty}  
\title{Kolmogorov–Arnold Networks for Personal Context Recognition on ExtraSensory}
%
%
\author{
Hoang-Thang Ta\inst{1,2}\orcidID{0000-0003-0321-5106}
} 
%
\authorrunning{Ta et al.}
%
\institute{
University of Information Technology, Ho Chi Minh City, Vietnam\\
\and
Vietnam National University Ho Chi Minh City, Ho Chi Minh City, Vietnam\\
\email{thangth@uit.edu.vn}
}
%
\maketitle              
\begin{abstract}
Kolmogorov--Arnold Networks (KANs) have attracted increasing attention in recent years, with applications across a wide range of AI tasks. In this paper, we evaluate several KAN variants on the ExtraSensory dataset for personal context recognition and compare them with a multilayer perceptron (MLP) and TabM. We conduct the main experiments using five user folds and three random seeds per fold and report the average Macro-F1, Micro-F1, and training time. We also perform shallow ablation studies on grid size, the number of grids, and data normalization to examine their effects on KAN performance. The results show that all evaluated KAN variants significantly outperform MLP in terms of Macro-F1 and Micro-F1 and achieve performance comparable to TabM. However, KAN variants generally require more training time, while TabM provides a more favorable balance between predictive performance and training efficiency. These results suggest that KANs are promising for personal context recognition, while their computational efficiency remains an important challenge. Our source code is publicly available at: \url{https://github.com/hoangthangta/ExtraSensory-KANs}.

\end{abstract}
\section{Introduction}

Personal context recognition aims to infer a person's activities and surrounding context from mobile and wearable sensor data. Neural networks have been widely used to model complex patterns in high-dimensional sensor measurements, with MLPs providing a general-purpose architecture for such data~\cite{majidzadeh2021human,ojiako2023mlps}. More recently, KANs have been explored for human activity recognition (HAR) and wearable sensing, including incremental learning with heterogeneous wearable-sensor inputs~\cite{liu2024ikan}, continual online and on-device HAR~\cite{liu2026cool}, and wearable-sensor-based fall detection and activity recognition~\cite{jiang2026mdwd}. These studies highlight the potential of KAN-based approaches for sensor-driven recognition tasks, while their application to broader personal context recognition remains limited.

The ExtraSensory dataset provides a challenging benchmark characterized by heterogeneous mobile and wearable sensor data and diverse contextual labels. Previous studies have applied LSTM- and RNN-based methods, together with feature selection and optimization techniques, to this benchmark~\cite{khan2024wearable,al2025iot}. Its multi-label setting, in which multiple contextual states can occur simultaneously, provides a useful testbed for evaluating different neural architectures beyond conventional sequence models.

In this work, we investigate KAN-based approaches for multi-label personal context recognition using ExtraSensory. We evaluate several representative KAN variants under a common experimental framework, comparing them with MLP as a conventional neural-network baseline and TabM as a recent tabular deep learning architecture based on parameter-efficient ensembling~\cite{gorishniy2025tabm}. We do not consider CNNs or RNNs because they are primarily designed to exploit spatial and temporal structure, whereas ExtraSensory provides fixed-dimensional, engineered features rather than raw sensor streams. We further conduct shallow ablation studies on grid size and data normalization and analyze training time to assess both predictive performance and computational efficiency. The main contribution is an empirical benchmark of KAN-based approaches for this task, providing a consistent comparison with MLP and TabM while examining the effects of grid size and normalization.

\section{Related Work} 

KANs were recently introduced by \citet{liu2025kan} as a new neural network architecture. Their design is motivated by the Kolmogorov--Arnold representation theorem (KART), which states that a continuous multivariate function can be represented as a finite combination of univariate functions~\cite{kolmogorov1957representation,braun2009constructive}. This theorem had already inspired several neural network architectures before the introduction of KANs, including spline-based models~\cite{leni2013kolmogorov,van2022kasam}. Since their introduction, KANs have been applied to a broad range of tasks, including differential equation solving~\cite{wang2025kolmogorov}, time-series forecasting~\cite{genet2024temporal}, and computer vision~\cite{li2025u}. Meanwhile, numerous studies have explored alternative representations for the learnable univariate functions, including polynomial functions~\cite{ss2024chebyshev}, radial basis functions~\cite{li2024kolmogorov}, Fourier bases~\cite{xu2024fourierkan}, wavelets~\cite{bozorgasl2024wav}, rational and fractional Jacobi functions~\cite{aghaei2026rkan,aghaei2025fkan}, and customized activation functions~\cite{qiu2025relu}.

Many studies have explored the use of KANs for HAR and wearable sensing. iKAN introduces KAN-based incremental learning for heterogeneous wearable-sensor inputs~\cite{liu2024ikan}, while COOL extends KANs to continual online and on-device HAR~\cite{liu2026cool}. MDWD-KAN further incorporates multilevel discrete wavelet decomposition into KANs for wearable-sensor-based fall detection and activity recognition~\cite{jiang2026mdwd}. In parallel, several studies have investigated conventional deep-learning approaches on ExtraSensory. These approaches combine signal filtering, feature extraction, recursive feature elimination (RFE), SMOTE, and LSTM for activity and localization recognition~\cite{khan2024wearable}; employ Boruta feature selection and particle swarm optimization (PSO) with parallel RNNs for activity recognition and localization~\cite{al2025iot}; and use a hybrid residual CNN and spatial LSTM architecture to process time--frequency representations of wearable-sensor signals for complex activity recognition~\cite{tian2024complex}.

Despite these developments, the application of KANs to broader personal context recognition remains largely unexplored. In particular, to the best of our knowledge, no existing study has evaluated KANs on the ExtraSensory dataset~\cite{vaizman2017recognizing}, which provides mobile and wearable sensing data for recognizing a diverse range of personal activities and contexts. This gap motivates our study, which investigates KANs for personal context recognition on ExtraSensory and compares them with MLP and TabM~\cite{gorishniy2025tabm}, a tabular architecture based on efficient ensembling of multiple parameter-sharing MLPs.

\section{Methodology}

\subsection{MLP}

MLP is a standard feed-forward neural network consisting of fully connected layers with nonlinear activation functions. For layer $l$, the MLP defines a mapping $f_l:\mathbb{R}^{n_l}\rightarrow\mathbb{R}^{n_{l+1}}$ as
\begin{equation}
f_l(\mathbf{x}_l)=\sigma(\mathbf{W}_l\mathbf{x}_l+\mathbf{b}_l),
\label{eq:mlp_layer}
\end{equation}
where $\mathbf{W}_l$ and $\mathbf{b}_l$ are the learnable weight matrix and bias vector, respectively. For an MLP with $L$ layers, the complete network is the composition
\begin{equation}
\mathrm{MLP}(\mathbf{x})=(f_{L-1}\circ\cdots\circ f_1\circ f_0)(\mathbf{x}).
\label{eq:mlp}
\end{equation}
In this work, MLP serves as a conventional baseline for comparison with KAN variants and TabM.



\subsection{TabM}

TabM is a tabular deep learning architecture that extends the multilayer perceptron (MLP) with parameter-efficient ensembling~\cite{gorishniy2025tabm}. It uses $k$ implicit MLP submodels trained in parallel that share most of their weights, while lightweight non-shared multiplicative and additive adapters provide diversity. The final prediction is obtained by averaging the predictions of the $k$ submodels, with a default ensemble size of $k=32$.

For an input $\mathbf{x}$, let $f_j(\mathbf{x})$ denote the prediction of the $j$-th implicit submodel. The individual predictions are given by
\begin{equation}
\hat{\mathbf{y}}_j = f_j(\mathbf{x}), \qquad j=1,\ldots,k,
\end{equation}
and the final prediction is computed as
\begin{equation}
\hat{\mathbf{y}}
=
\frac{1}{k}\sum_{j=1}^{k}\hat{\mathbf{y}}_j.
\end{equation}
This parameter-sharing strategy reduces the parameter overhead compared with a conventional ensemble of independently trained MLPs while retaining the diversity associated with multiple prediction branches. The original study also shows that the collective prediction of the implicit submodels can provide better generalization than their individual predictions~\cite{gorishniy2025tabm}.

\subsection{KAN}

KANs are inspired by the KART, which states that any continuous multivariate function on a bounded domain can be expressed as a finite combination of one-dimensional functions~\cite{kolmogorov1957representation}. Based on this theorem, \citet{liu2025kan} introduced KANs as a new neural network architecture. In contrast to MLPs, which use fixed activation functions and learnable linear weights, KANs use learnable one-dimensional functions on the edges. A KAN with $L$ layers applies a sequence of function matrices $\Phi_0, \Phi_1, \ldots, \Phi_{L-1}$:

\begin{equation}
\begin{aligned}
\text{KAN}(\mathbf{x}) = (\Phi_{L-1} \circ \cdots \circ \Phi_0)(\mathbf{x})
\end{aligned}
\label{eq:kan}
\end{equation}

For layer $l$, the function matrix $\Phi_l \in \mathbb{R}^{n_{l+1}\times n_l}$ consists of learnable functions $\phi_{l,j,i}$, where each function connects the $i$-th node in layer $l$ to the $j$-th node in layer $l+1$:

\begin{equation}
\begin{aligned}
\phi_{l,j,i}, \quad l = 0, \cdots, L - 1, \quad i = 1, \cdots, n_l, \quad j = 1, \cdots, n_{l+1}
\end{aligned}
\label{eq:acti_funct}
\end{equation}
Each function matrix $\Phi_l$ comprises pre-activation functions $\phi_{l,j,i}$, where each function defines the connection from neuron $(l,i)$ to neuron $(l+1,j)$. The matrix $\Phi_l$ is written as:

\begin{equation}
\begin{aligned}
\mathbf{x}_{l+1} = 
\underbrace{\left(
\begin{array}{cccc}
\phi_{l,1,1}(\cdot) & \phi_{l,1,2}(\cdot) & \cdots & \phi_{l,1,n_l}(\cdot) \\
\phi_{l,2,1}(\cdot) & \phi_{l,2,2}(\cdot) & \cdots & \phi_{l,2,n_l}(\cdot) \\
\vdots & \vdots & \ddots & \vdots \\
\phi_{l,n_{l+1},1}(\cdot) & \phi_{l,n_{l+1},2}(\cdot) & \cdots & \phi_{l,n_{l+1},n_l}(\cdot)
\end{array}\right)}_{\Phi_{l}} \mathbf{x}_l
\label{eq:function_matrix}
\end{aligned}
\end{equation}

\begin{figure*}[!htbp]
  \centering
\includegraphics[scale=0.7]{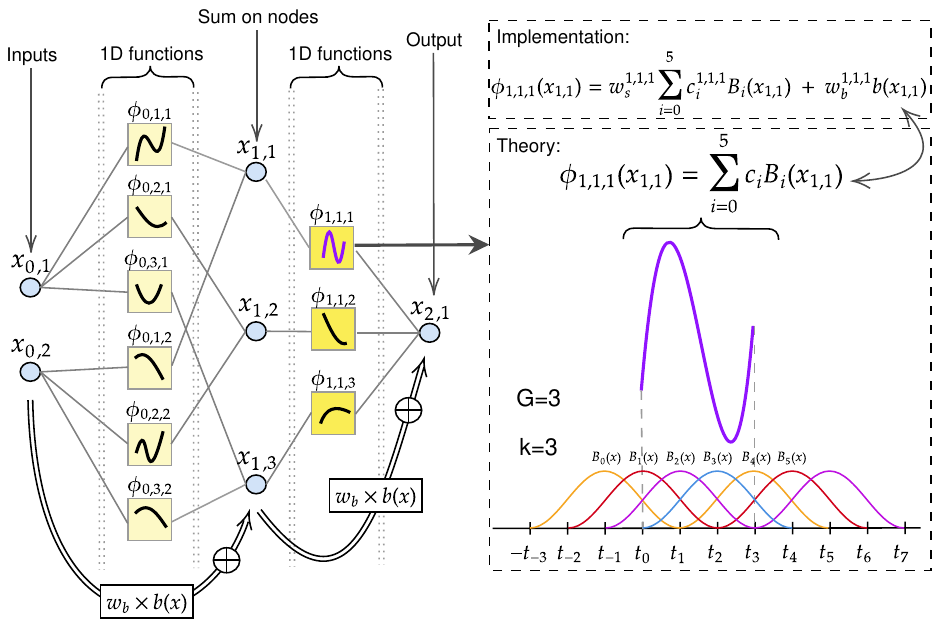}
  \centering
  \caption{Left: The architecture of KAN(2,3,1). Right: Calculation of $\phi_{1,1,1}$ using control points and B-spline basis functions, where $G$ and $k$ denote the grid size and spline order, respectively, and $n=G+k=3+3=6$ is the number of B-spline basis functions. The figure is reused from~\citet{ta2026sechkan}.}
\label{fig:kan_diagram}
\end{figure*}

In the original KAN, each learnable function is defined as a combination of a base function and a spline function~\cite{liu2025kan}:

\begin{equation}
\begin{aligned}
\phi(x) = w_b b(x) + w_s spline(x)
\end{aligned}
\label{eq:acti_funct_imp}
\end{equation}
The base function $b(x)$ is the SiLU activation function:

\begin{equation}
\begin{aligned}
b(x) = silu(x) = \frac{x}{1 + e^{-x}}
\end{aligned}
\label{eq:b_function}
\end{equation}
The spline component is represented as a weighted sum of B-spline basis functions:

\begin{equation}
\begin{aligned}
spline(x) = \sum_{i}c_iB_i(x)
\end{aligned}
\label{eq:spline_function}
\end{equation}
where $c_i$ are learnable coefficients and $B_i(x)$ denote the B-spline basis functions.

\Cref{fig:kan_diagram} illustrates the architecture of KAN(2,3,1), which consists of 2 input nodes, 3 hidden nodes, and 1 output node. Each node computes its output by summing the functions $\phi$ associated with its incoming edges. The figure also shows the construction of $\phi$ using control points and B-spline basis functions. With $G=3$ and spline order $k=3$, the number of B-splines is $G+k=3+3=6$, and the index $i$ therefore ranges from 0 to 5. The theoretical formulation of $\phi$ differs from its implementation in KAN (\Cref{eq:acti_funct_imp}), where the spline and base components are combined using the spline weight matrix $w_s$ and base weight matrix $w_b$, with the base activation function $b(x)$ applied to the input $x$.

\begin{table*}[htbp]
\centering
\caption{Summary of KAN variants and their basis functions.}
\label{tab:kan_variants}
\begin{tabular}{lp{8cm}}
\hline
\textbf{Variant} & \textbf{Basis Function} \\ \hline

EfficientKAN~\cite{Blealtan2024} &
$\phi(x)=w_b b(x)+w_s B(x)$, where
$B(x)=\sum_i c_iB_i(x)$, with $c_i$ denoting spline coefficients and
$B_i(x)$ denoting B-spline basis functions.
\\ \hline

FastKAN~\cite{li2024kolmogorov} &
$\phi(x)=w_b b(x)+w_s\mathit{RBF}(x)$, where
$\mathit{RBF}(x)=\exp\!\left(-\frac{\|x-d\|^2}{2h^2}\right)$;
$d$ is the center and $h$ is the function width or spread.
\\ \hline

BSRBF-KAN~\cite{ta2024bsrbf} &
$\phi(x)=w_b b(x)+w_s\big(\mathit{B}(x)+\mathit{RBF}(x)\big)$,
where $\mathit{B}(x)$ and $\mathit{RBF}(x)$ are defined above.
\\ \hline

\end{tabular}
\end{table*}

KANs have motivated a range of variants that modify the univariate function representation to enhance computational efficiency and modeling flexibility. In this work, we consider several popular KANs in our experiments. EfficientKAN~\cite{Blealtan2024} formulates the univariate function using a base function together with a B-spline expansion, controlled by learnable scaling factors. FastKAN~\cite{li2024kolmogorov} replaces the B-spline representation with Gaussian radial basis functions (RBFs), resulting in a more efficient functional representation. BSRBF-KAN~\cite{ta2024bsrbf} combines B-spline and Gaussian RBF components within a unified representation to leverage their complementary characteristics. The basis functions of these KANs are summarized in \Cref{tab:kan_variants}.

\section{Experiments}

\subsection{Datasets and Training Configurations}
\begin{table*}[!ht]
	\caption{Network structures and numbers of trainable parameters for the models on the ExtraSensory dataset.}
	\centering
	\resizebox{\textwidth}{!}{%
	\begin{tabular}{p{2.2cm}p{1.5cm}p{8.0cm}}
            \hline
		\textbf{Model} & \textbf{Params} & \textbf{Network structure} \\
            \hline
		BSRBF-KAN 
		& 72,544 
		& $225$--$29$--$51$; grid size $=5$, spline order $=3$, LayerNorm, SiLU. \\
		
		EfficientKAN 
		& 71,760 
		& $225$--$26$--$51$; grid size $=5$, spline order $=3$, SiLU. \\
		
		FastKAN 
		& 72,624 
		& $225$--$29$--$51$; grid size $=8$, LayerNorm, SiLU. \\
		
		MLP 
		& 71,925 
		& $225$--$256$--$51$; LayerNorm and SiLU. \\
		
		TabM 
		& 71,002 
		& 225 numerical features; 1 block, $k=8$. \\
            \hline
		\multicolumn{3}{l}{Params = number of trainable parameters.} \\
            \hline
	\end{tabular}%
	}
	\label{tab:model_architectures}
\end{table*}

We evaluate the models on ExtraSensory, a real-world personal context recognition dataset containing 225 input features and 51 labels collected from 60 users. Each observation is represented by a 225-dimensional feature vector, and the multi-label classification task predicts a 51-dimensional binary label vector, with multiple contextual labels potentially associated with the same observation. We use the 51 cleaned labels provided with the ExtraSensory dataset. Following the official five-fold user partition, each fold contains 12 test users. For each fold, we further split the remaining 48 users into 38 training users and 10 validation users using an approximately 80/20 subject-level split with a random seed of 42. Missing feature values are handled using training-set statistics, while unavailable labels are excluded using the corresponding label masks during both training and evaluation.

For multi-label evaluation, Macro-F1 is computed as the mean of the per-label F1 scores over labels with at least one observed positive instance, whereas Micro-F1 is computed by aggregating TP, FP, and FN across all observed label entries. For checkpoint selection, we use balanced accuracy, computed as the mean of the per-label balanced accuracies over labels with at least one observed positive and one observed negative instance. For each label, balanced accuracy is defined as \((\mathrm{TPR}+\mathrm{TNR})/2\). Model predictions are obtained by applying a sigmoid function to the output logits and using a threshold of 0.5 to obtain binary predictions.

We compare MLP, EfficientKAN, FastKAN, BSRBF-KAN, and TabM under a common training configuration. All models are trained for 20 epochs with a batch size of 64 using AdamW with a learning rate of $10^{-3}$ and weight decay of $10^{-4}$, together with a OneCycleLR scheduler to facilitate effective optimization within the limited number of training epochs. Class balancing is applied during training, and the best checkpoint is selected based on validation balanced accuracy. The selected checkpoint is then evaluated on the held-out test users.

MLP uses a 256-unit hidden layer with SiLU activation and LayerNorm. EfficientKAN uses 26 hidden units with SiLU activation and no additional normalization, while FastKAN uses 29 hidden units with SiLU activation, LayerNorm, and a number of grids of 8 (\texttt{num\_grids=8}). BSRBF-KAN uses 29 hidden units with SiLU base activation, LayerNorm, a grid size of 5 (\texttt{grid\_size=5}), and spline order 3 (\texttt{spline\_order=3}). To ensure a fair comparison among the KAN variants, we use commonly adopted values for the grid configuration. Specifically, for FastKAN, we set \texttt{num\_grids = grid\_size + spline\_order}. TabM is configured with a single block and $k=8$ ensemble members. The resulting network structures and numbers of trainable parameters are summarized in \Cref{tab:model_architectures}. Each configuration is evaluated using three random seeds (0, 1, and 2) across the five folds, and the results are reported as mean $\pm$ standard deviation over 15 runs.

\subsection{Results}

\begin{table}[ht]
\centering
\caption{Multi-label classification results on ExtraSensory. Each model is evaluated over 15 runs (5 folds and 3 random seeds). Results are reported as mean $\pm$ standard deviation, and training time is reported as the average over those runs.}
\label{tab:extrasensory_multilabel_results}
\begin{tabular}{p{2.5cm}p{2.8cm}p{2.8cm}p{3cm}}
\hline
\textbf{Model} & \textbf{Macro-F1} & \textbf{Micro-F1} & \textbf{Train Time (s)} \\
\hline
BSRBF-KAN    & $0.3106 \pm 0.0095$ & $\mathbf{0.5572 \pm 0.0123}$ & 665.47 \\
EfficientKAN & $\mathbf{0.3112 \pm 0.0072}$ & $0.5556 \pm 0.0179$ & 540.80 \\
FastKAN      & $0.3091 \pm 0.0111$ & $0.5527 \pm 0.0143$ & 463.80 \\
MLP          & $0.2807 \pm 0.0099$ & $0.4957 \pm 0.0184$ & $\mathbf{372.86}$ \\
TabM         & $0.3107 \pm 0.0089$ & $0.5562 \pm 0.0087$ & 381.30 \\
\hline
\end{tabular}
\label{tab:main_results}
\end{table}

\begin{table}[!ht]
\centering
\caption{Pairwise Wilcoxon signed-rank tests for multi-label classification.
The tests are conducted over 15 matched fold--seed runs, and $p$-values are adjusted using Holm correction.}
\label{tab:wilcoxon_test}
\begin{tabular}{p{2cm}p{5cm}p{1.5cm}p{1.5cm}p{1.5cm}}
\hline
\textbf{Metric} & \textbf{Comparison} & \textbf{$p$} & \textbf{$p_{\mathrm{adj}}$} & \textbf{Sig.} \\
\hline
Macro-F1 & BSRBF-KAN vs.\ EfficientKAN & 0.6387 & 1.0000 & No \\
         & BSRBF-KAN vs.\ FastKAN      & 0.5995 & 1.0000 & No \\
         & BSRBF-KAN vs.\ MLP          & 0.0001 & 0.0012 & Yes \\
         & BSRBF-KAN vs.\ TabM         & 0.8469 & 1.0000 & No \\
         & EfficientKAN vs.\ FastKAN   & 0.5995 & 1.0000 & No \\
         & EfficientKAN vs.\ MLP       & 0.0001 & 0.0012 & Yes \\
         & EfficientKAN vs.\ TabM      & 0.7197 & 1.0000 & No \\
         & FastKAN vs.\ MLP            & 0.0001 & 0.0012 & Yes \\
         & FastKAN vs.\ TabM           & 0.3303 & 1.0000 & No \\
         & MLP vs.\ TabM               & 0.0001 & 0.0012 & Yes \\
\hline
Micro-F1 & BSRBF-KAN vs.\ EfficientKAN & 0.8040 & 1.0000 & No \\
         & BSRBF-KAN vs.\ FastKAN      & 0.2293 & 1.0000 & No \\
         & BSRBF-KAN vs.\ MLP          & 0.0001 & 0.0012 & Yes \\
         & BSRBF-KAN vs.\ TabM         & 0.8040 & 1.0000 & No \\
         & EfficientKAN vs.\ FastKAN   & 0.5245 & 1.0000 & No \\
         & EfficientKAN vs.\ MLP       & 0.0001 & 0.0012 & Yes \\
         & EfficientKAN vs.\ TabM      & 0.6788 & 1.0000 & No \\
         & FastKAN vs.\ MLP            & 0.0001 & 0.0012 & Yes \\
         & FastKAN vs.\ TabM           & 0.5995 & 1.0000 & No \\
         & MLP vs.\ TabM               & 0.0001 & 0.0012 & Yes \\
\hline
\multicolumn{5}{l}{\footnotesize
$p$ = raw $p$-value; $p_{\mathrm{adj}}$ = Holm-adjusted $p$-value; }
\\
\multicolumn{5}{l}{\footnotesize
Sig. = significant at $p_{\mathrm{adj}} < 0.05$.}
\\
\hline
\end{tabular}
\end{table}

As shown in \Cref{tab:main_results}, KAN variants consistently achieve higher Macro-F1 and Micro-F1 than MLP, while TabM achieves performance comparable to the KAN-based models. EfficientKAN obtains the highest Macro-F1, whereas BSRBF-KAN obtains the highest Micro-F1. However, the pairwise Wilcoxon signed-rank tests in \Cref{tab:wilcoxon_test} show no statistically significant differences among the KAN variants and TabM after Holm correction. In contrast, all KAN variants and TabM significantly outperform MLP on both Macro-F1 and Micro-F1. Relative to MLP, TabM improves Macro-F1 by approximately 10.7\% and Micro-F1 by approximately 12.2\%, while increasing training time by only about 2.3\%. Thus, the higher predictive performance of these models does not require a proportional increase in training time.

The training-time results further reveal differences in computational efficiency. Among the KAN variants, FastKAN has the shortest training time, requiring approximately 14.3\% less time than EfficientKAN and 30.3\% less than BSRBF-KAN, while achieving similar predictive performance. TabM requires approximately 29.5\% less training time than EfficientKAN and 17.8\% less than FastKAN, while maintaining comparable Macro-F1 and Micro-F1. MLP remains the fastest model, but its lower predictive performance is accompanied by only a small training-time advantage over TabM. These results indicate that TabM provides a competitive balance between predictive performance and training efficiency, while FastKAN offers the lowest training cost among the evaluated KAN variants.

\subsection{Ablation studies}

\begin{figure*}[!ht]
  \centering
  \includegraphics[scale=0.50]{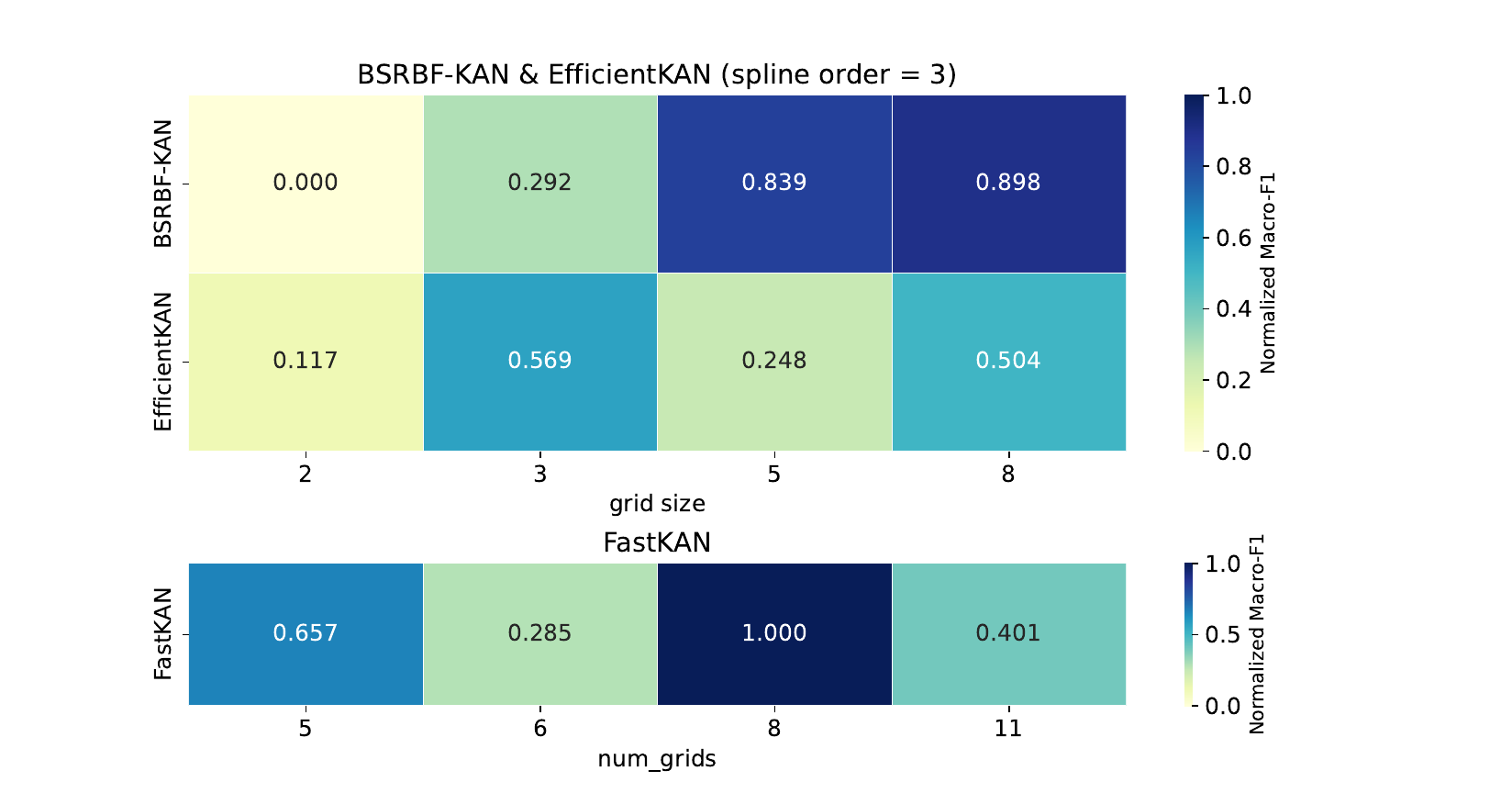}
  \caption{Ablation study of grid size and the number of grids (\texttt{num\_grids}) in BSRBF-KAN, EfficientKAN, and FastKAN. Each model was evaluated on fold 0 using three random seeds (0, 1, and 2). The average Macro-F1 across the three seeds was normalized to the range $[0,1]$ for visualization in the heatmap, where 0 and 1 denote the lowest and highest results, respectively. For BSRBF-KAN and EfficientKAN, \texttt{spline\_order = 3} and \texttt{grid\_size} $\in \{2,3,5,8\}$. For FastKAN, \texttt{num\_grids} $\in \{5,6,8,11\}$, corresponding to \texttt{grid\_size} $\in \{2,3,5,8\}$ under \texttt{num\_grids = grid\_size + spline\_order}.}
  \label{fig:ab_grid}
\end{figure*}

\Cref{fig:ab_grid} presents the ablation results for different grid sizes and numbers of grids in BSRBF-KAN, EfficientKAN, and FastKAN. To limit computational cost, the ablation is conducted on fold 0 using three random seeds (0, 1, and 2). The heatmap reports the average Macro-F1 over three random seeds (0, 1, and 2), normalized to $[0,1]$ within each model for visualization. As shown in \Cref{fig:ab_grid}, BSRBF-KAN achieves its highest normalized Macro-F1 at \texttt{grid\_size = 8}, followed by \texttt{grid\_size = 5} and 3, while \texttt{grid\_size = 2} performs worst. For EfficientKAN, the best result is obtained at \texttt{grid\_size = 3}, followed by 8, 5, and 2. For FastKAN, we vary \texttt{num\_grids} over $\{5,6,8,11\}$, corresponding to \texttt{grid\_size} values of $\{2,3,5,8\}$ under \texttt{num\_grids = grid\_size + spline\_order} with \texttt{spline\_order = 3}. The highest normalized Macro-F1 is achieved with \texttt{num\_grids = 8} (\texttt{grid\_size = 5}), followed by \texttt{num\_grids = 5}, 11, and 6. Generally, the results do not show a monotonic relationship between grid size and Macro-F1, indicating that increasing the grid size does not necessarily lead to better performance.

The main experiments use the predefined architectures and normalization configurations in \Cref{tab:extrasensory_multilabel_results}. The ablation studies are conducted separately and are not used to retune the main experimental configurations. We further conduct an ablation study on data normalization strategies. BSRBF-KAN supports BatchNorm, LayerNorm, and No Norm, while FastKAN supports LayerNorm and No Norm. For EfficientKAN, we retain its original architecture, which does not use a normalization layer. As shown in \Cref{tab:ab_norm}, BSRBF-KAN achieves its highest Macro-F1 with BatchNorm, followed by LayerNorm and No Norm. Similarly, FastKAN achieves a higher Macro-F1 with LayerNorm than without normalization. These results show that the normalization strategy affects Macro-F1, with normalization providing a higher average Macro-F1 for both BSRBF-KAN and FastKAN in this experiment.

\begin{table}[!ht]
\centering
\caption{Macro-F1 results for various normalization strategies on ExtraSensory. Results are averaged over three seeds and reported as mean $\pm$ standard deviation.}
\label{tab:ab_norm}
\begin{tabular}{p{3.5cm}p{4cm}p{3cm}}
\hline
\textbf{Model} & \textbf{Normalization} & \textbf{Macro-F1} \\
\hline
BSRBF-KAN & BatchNorm & $\mathbf{0.3214 \pm 0.0058}$ \\
           & LayerNorm & $0.3168 \pm 0.0177$ \\
           & No Norm   & $0.3153 \pm 0.0113$ \\
\hline
FastKAN   & LayerNorm & $\mathbf{0.3190 \pm 0.0013}$ \\
           & No Norm   & $0.3089 \pm 0.0051$ \\
\hline
\end{tabular}
\end{table}

\section{Limitations}
Although this study obtains some positive results, it still has some limitations. First, we evaluate the models only on ExtraSensory and do not consider other personal context recognition datasets. Second, we do not compare KANs with other neural networks commonly used for sensor-based recognition, such as CNNs and LSTMs. Third, we do not analyze the characteristics of ExtraSensory's data to understand why KANs perform differently on this dataset. Finally, we do not perform an extensive hyperparameter search for the KAN variants. Additional experiments using more datasets, models, data analyses, and hyperparameter settings can help address these limitations.

\section{Conclusion}
In this paper, we investigated several popular KAN variants for personal context recognition on ExtraSensory. We compared them with MLP and TabM under a common multi-label classification setting using five user folds and three random seeds per fold. The results show that all evaluated KAN variants significantly outperform MLP in terms of Macro-F1 and Micro-F1, while achieving performance comparable to TabM. Among the KAN variants, EfficientKAN achieves the highest Macro-F1, while BSRBF-KAN achieves the highest Micro-F1. However, the KAN variants require more training time than TabM, while FastKAN provides the lowest training cost among the evaluated KAN variants.
These results suggest that KANs are promising for personal context recognition, but their computational efficiency remains an important challenge. TabM provides a competitive balance between predictive performance and training time in the current experimental setting. Future work will evaluate KANs on additional personal context recognition datasets, compare them with other neural network architectures, investigate the characteristics of ExtraSensory that may explain the observed performance, and perform more extensive hyperparameter optimization. These additional experiments can provide a better understanding of the strengths and limitations of KANs for sensor-based personal context recognition.

\section*{Acknowledgements}
This research was funded by the University of Information Technology, Vietnam National University Ho Chi Minh City, under Grant No. S4-2027-80616.

%
%
%
\bibliographystyle{splncs04nat}
\bibliography{references}
\end{document}